\documentclass[10pt,a4paper,twocolumn]{article}
\usepackage{Style5}
\usepackage{url,hyperref}
\usepackage[switch]{lineno}

\usepackage{lipsum}
\usepackage{graphicx}
\begin{document}
\title{Evolution is Driven by Natural \\ Autoencoding: Reframing Species,  \\ Interaction Codes, 
Cooperation, and Sexual Reproduction.}

\author[1]{Irun R. Cohen{$^*$}}
\author[2]{Assaf Marron}
\affil[1]{Department of Immunology and Regenerative Biology,   Weizmann Institute of Science, Rehovot, 76100 Israel, irun.cohen@weizmann.ac.il}
\affil[2]{Department of Computer Science and Applied Mathematics, Weizmann Institute of Science, Rehovot, 76100 Israel, assaf.marron@weizmann.ac.il}

\maketitle

\begin{abstract}
The continuity of life and its evolution, we proposed, emerge from an interactive group process manifested in networks of interaction. We term this process \textit{survival-of-the-fitted}.
Here, we reason that survival of the fitted results from a natural computational process we term \textit{natural autoencoding}. Natural autoencoding works by retaining repeating biological interactions while non-repeatable interactions disappear.\\ 
1) We define a species by its \textit{species interaction code}, which  consists of  a compact description of the repeating interactions of species organisms with their external and internal environments. Species interaction codes are descriptions recorded in the biological infrastructure  that enables repeating interactions.  Encoding and decoding are interwoven.\\ 
2) Evolution proceeds by natural autoencoding of sustained changes in species interaction codes. DNA is only one element in natural autoencoding.\\ 
3) Natural autoencoding accounts for the paradox of genome randomization in sexual reproduction---recombined genomes are analogous to the diversified inputs required for artificial autoencoding.
The increase in entropy generated by genome randomization compensates for the decrease in entropy generated by organized life.\\ 
4) Natural autoencoding and artificial autoencoding algorithms manifest defined similarities and differences.\\
Recognition of the importance of fittedness could well serve the future of a humanly livable biosphere.   
\end{abstract}

\clearpage 

\section*{Keywords}

Interaction, Autoencoding, Species Interaction Code, Sexual Reproduction, Survival of the Fitted

\section{Background and Aims}\label{sec:BackgroundAims}

Previously, we proposed: 

(1) that interactions are the vehicle of biologic
evolution---what evolves are the interactions of entities;

(2) that cooperative group
interaction networks are more functional in evolution than are individual competition and survival of the fittest---the outcome 
of evolution is not survival of only the reproductively dominant individuals
but survival of integrated group networks; we have termed this outcome \textit{survival-of-the-fitted} 
~\cite{cohenMarron2020survivalOfTheFitted,cohen2000tendingAdamsGarden,cohen2016updatingDarwin};
and

\begin{figure*}[t] 
	\centering
	\includegraphics[width=0.95\linewidth]{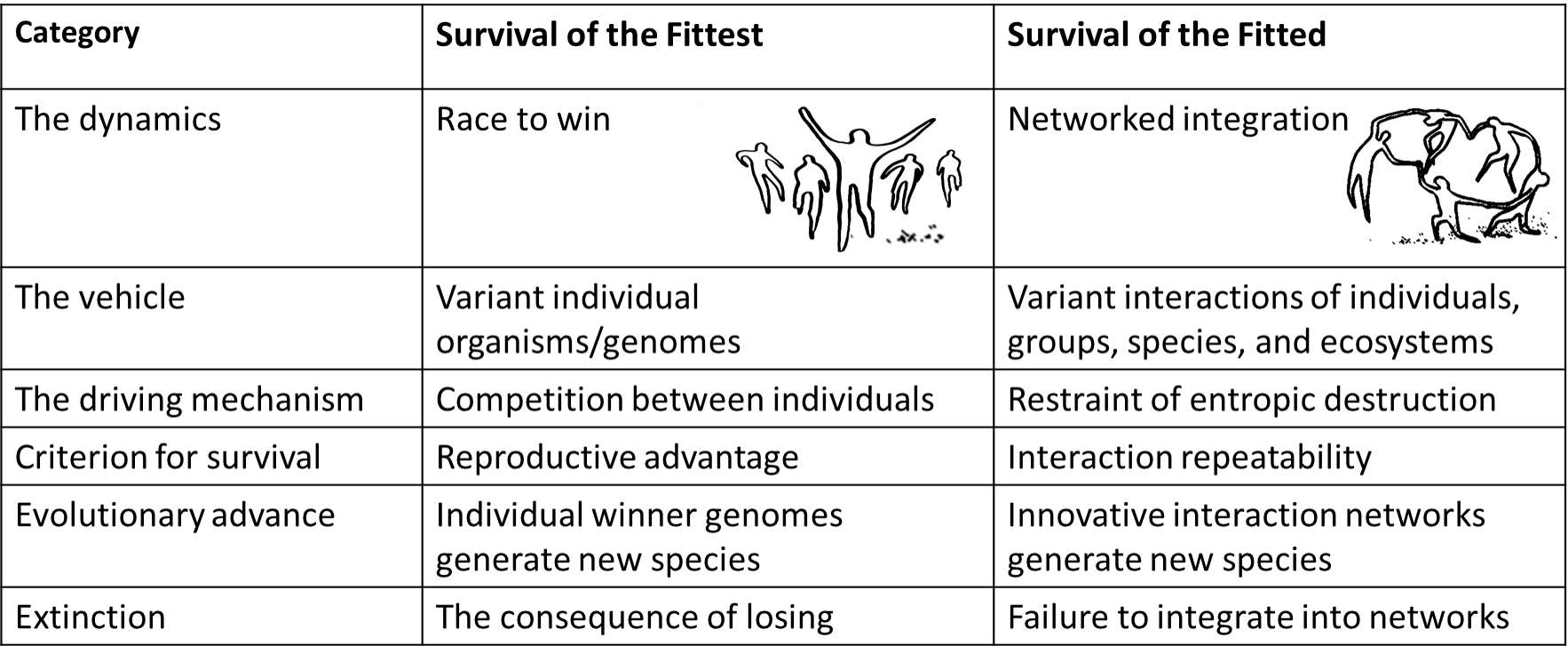}
	\caption{Two mechanisms of evolution:  Differences between \textit{Survival of the Fittest} and \textit{Survival of the Fitted}. (Image credits: Henri Matisse; Rachel Shiloach)}
	\label{fig:fittestFitted}
	
\end{figure*}

(3) that evolution takes place in accord with the laws of physical nature, including the dissolution of order dictated 
by 
the second law of thermodynamics and the continuous increase of entropy~\cite{cohenMarron2020survivalOfTheFitted}. 
Figure~\ref{fig:fittestFitted} summarizes salient differences between survival of the fittest and survival of the fitted. 

This paper extends these ideas and supports two conclusions:  first, that species can be defined 
as descriptions of codes of interaction common to the collective of species organisms, 
and secondly that a computational enterprise we term \textit{natural autoencoding} generates the evolution of species and of the biosphere.
Below, we define some of the key terms we use in developing these ideas. 

\section{Basic Definitions}\label{sec:defs}

\subsubsection*{Evolution}

 Evolution is the narrative of changes in species and their interactions over time~\cite{koonin2011logicOfChance}.

\subsubsection*{Environment}

The environment is the aggregate of the living and non-living entities, structural and dynamic, within which an organism exists and operates.

\subsubsection*{Survival-of-the-Fittest; Natural Selection}
Survival-of-the-fittest is a term used since Darwin to describe the mechanism that drives evolution~\cite{darwin1860originOfSpecies}; Survival-of-the-fittest assumes a continuous struggle of variant individual organisms for survival and reproductive advantage in the face of limited resources;
this struggle leads to survival of the \textit{fittest}
individuals and their dominion over
less fit individuals; 
predominating proliferation of the winners determines the characters of species.  This process of reward is termed \textit{natural selection}. Variants of natural selection have been proposed to account for advantages in cooperation as well as in struggle~\cite{nowak2006fiveRulesEvolutionOfCooperation}.

\subsubsection*{Survival-of-the-Fitted}

 It is now clear that all living systems---cells, organisms, species (including \textit{Homo sapiens}) 
and ecosystems---survive in
extensive networks of interaction and group
cooperation~\cite{cohen2016updatingDarwin,cohenMarron2020survivalOfTheFitted,sachs2004evolutionOfCoop}.
A few examples include the dependence of every multi-cellular organism on a resident microbiome~\cite{blaser2014microbiomeRevolution}; the symbiotic web of forest trees and fungi~\cite{simard2018mycorrhizal}; and the collaboration and symbiosis  that create a coral colony~\cite{rosenberg2007coralSymbiosis}. 
The biosphere is sustained by such repeatable interactions;
the biosphere is a \emph{worldwide web} of interactions.

Survival-of-the-fitted is an alternative mechanism to account for evolution~\cite{cohenMarron2020survivalOfTheFitted,cohen2016updatingDarwin,cohen2000tendingAdamsGarden}.
Survival does not result only from individual struggles for reproductive advantage; surviving organisms are those that integrate into networks of sustaining interactions; 
longevity and 
rates of reproduction are not individual achievements but to a large extent are expressed in the lifestyle of the  
species---some species of organisms live and reproduce for a season, some for a century, or more; some produce many offspring and some few. 

Survival-of-the-fitted affirms that what works, works~\cite{cohenMarron2020survivalOfTheFitted}.
By contrast, Darwinian survival-of-the-fittest would claim that what wins, works.  

\subsubsection*{Interaction}

 Interactions are mutual relationships between two or more entities
(termed interactors) in which the interactors 
transmit or exchange 
energy, matter or information; an exchange of information often involves matter and energy.
A \textit{process} is an ordered set of interactions.

Sustainable interactions are characterized by \textit{repetition and sequence}.
Metabolic interactions, for example, are organized in repeating, sequential pathways---in each pathway one interaction is connected to the next in line~\cite{judgeDodd2020Metabolism}.

Cycles of reproduction, growth, aging, illness, predation and death are accessible examples of the universality of repeated, sequential interactions.

\subsubsection*{Information and Meaning}

 We define information according to Shannon as a particular non-random structure or arrangement of entities or processes~\cite{shannon1948informationTheory,cohen2006informationallandscapesInArt}.
Arrangements bear information; but an arrangement by itself has no meaning unless it interacts with other arrangements 
to produce some effect~\cite{cohen2006informationallandscapesInArt,cohen2000tendingAdamsGarden}.
The consequences of the interactions of information constitute the \textit{meaning} of the information.
A sequence of DNA, for example, bears information that only gains meaning through expressed interactions including
transcription and translation~\cite{cohenAtlanEfroni2016geneticsAsExplanation}. Written words, too, have no meaning unless somebody or some thing can read them.  The meaning of information emerges from the information's interactions.

\subsubsection*{Energy and Matter}
 Energy, in functional terms, is the impetus behind motion and activity~\cite{doige2012typologyOfHeatAcrossTextbooks}, including the capacity to do work. Energy enables interactions.

Matter can be viewed as a product of interaction: the nuclei of atoms are created by interactions between fundamental particles; atoms are formed by interactions between nuclei and electrons; and molecules are formed by interactions between atoms.

So one can conclude that
anything made of atoms or molecules, including living entities and the biosphere itself, is made of interactions. As stated by Feynman~\cite[In Prologue]{gleick1993geniusFeynman} and others ~\cite{rovelli2018realityInteraction} interactions constitute reality.
This view is also in line with the relational philosophy of Leibniz and Whitehead~\cite{igamberdiev2018Leibniz},\cite[Ch.18]{routledge2009metaphysics}.

\subsubsection*{Code, Encoding and Decoding}

 The word \textit{code} can be defined in different ways~\cite{OED2022}.
The word  is derived from the Latin \textit{codex}, a book. We here define an \emph{interaction code} as a description (a ``book'') that outlines 
steps that, when implemented, are able to convert one form of biological information into another form of biological information.  
A biological code is analogous to the text of a computer algorithm that describes a set of repeatable interactions.

The term code can be used in at least four 
interwoven 
contexts.  
This is exemplified by the genetic code.

\begin{enumerate}
\item A concrete instance of input for a translation process: One concrete codon UUU is a code for generating molecules of phenylalanine.

\item A single decoding rule: \textit{``All codons UUU are codes for the molecules of amino acid phenylalanine''}, or \textit{``UUU maps to phenylalanine''}. 

\item A description of a set of reactive behaviors:
a DNA sequence that translates into a particular protein
is a code for all the reactive behaviors of this protein. Actually, the produced protein is another code for these reactions.

\item A complete system for such rules: this is exemplified by the very concept of the genetic code. Other such systems include the binary code used in computing to encode numbers, or the Dewey Decimal System for encoding book locations by subject in a library. Below we describe natural autoencoding, which is a process that forms code systems in nature. 
\end{enumerate}

A code may also feature information that summarizes or reduces to essentials the interactions that gave rise to the encoded information.
For example, a DNA code expresses the essence of the myriad of biological interactions---molecular, physiological and evolutionary---that have resulted in that sequence of DNA.
All of these many complicated interactions are reduced to the DNA sequence; this reduction encapsulates all the foregoing 
networks of interactions into a concise, manageable and functional code molecule, one that can be replicated and transmitted.

\subsubsection*{Encoding}
  The term \textit{encoding} refers to the process by which anterior interactions give rise to a derivative, often simplified new entity---the code. 
  The code, as we defined above, is a description;
encoding, unlike the derived code, 
is a process generated by actual interactions.

\subsubsection*{Decoding}

 \textit{Decoding}, like encoding, also is
a process---the interactions by which the encoded potential interactions get expressed---become actualized.
Decoding is a process that generates new information through interactions.  
 The distinction between description---code---and process---the encoding/decoding interactions---is useful;
in the course of our discussion, we shall relate this distinction to a natural autoencoding mechanism of evolution.

\subsubsection*{Natural autoencoding}
Below we shall analyze 
this concept in detail, but we introduce it here among our definitions.
Natural autoencoding is a process by which
repeating patterns of encoding and decoding, a complete code system, are formed and maintained. 

\section{Species Interaction Code}\label{sec:speciesInteractionCode}

 The concept of species, since Darwin, is linked to evolution; this link is reflected in the title of Darwin's foundational work: \textit{The Origin of Species}~\cite{darwin1860originOfSpecies}. 
A species originally referred to entities that look alike---the word \textit{species} derives from the Latin \textit{specere, to see}. 
Living species, basically, are composed of types of creatures that look alike,
and interact alike. 

The definition of a species beyond appearances is controversial:
A search in Google Scholar for \textit{species} returns millions of publications; but there is not one universally accepted definition;  researchers have proposed many 
different definitions of multi-cellular species based on morphology, genetics, sexual reproduction, ecology, and other criteria~\cite{mallet1995speciesDef}.

The definition of bacterial species is even more uncertain~\cite{chun2018prokaryoteSpecies} 
and we shall not deal in depth with prokaryotes or single-cell eukaryotes in this paper.
Unless designated otherwise, here the word species refers to multi-cellulars.

We define  a species as a collective of organisms
that jointly carry out 
a set of potential, repetitive interactions with their 
external and internal 
environment.

The code of each species is the ``book'' describing the essential interactions carried out by members of the species. 

Note that not every individual organism within a species need perform
all the coded interactions of the species---males, females, 
and particular ``sub-types'' of organisms within a species can perform uniquely different interactions that, nevertheless, are included in the collective species code. This is because the ensuing generations of species organisms, as a group,
continue to collectively fulfill the species  code, 
maintained, but not necessarily performed, by the reproducers.

Worker bees, for example, cannot themselves reproduce, but reproduction by queen bees and fertile males will continuously generate worker bees as part of the coded description of bees; likewise, the queen bee will never make honey, but the non-reproducing workers will.

In Section~\ref{sec:sexrep}
below, we discuss the function of sexual reproduction in maintaining and defining a multi-cellular species.  Organisms may exist in close connection with other organisms, but the species can be distinguished by independent sexual reproduction.  For example, every multi-cellular organism is accompanied by a resident microbiome, but the microbiome is not a member of the multi-cellular species; the microbiome organisms are not reproduced by the act of sexual reproduction of the multi-cellular organism---the microbiome is not part of the organism species, but has to be acquired independently.

The code of species interactions
comprises the information that, when decoded and expressed, enables the species to survive and thrive in the context of its environment.

The concepts of interaction code and species interaction code are different. 
Above in Section~\ref{sec:defs}, we defined a code for a given interaction, interaction code, as 
a description of the interaction. 
A species interaction code, however, has a broader meaning: a species interaction code is a description of \textit{sets} of interactions carried out at different times by the species collective;
a species interaction code is an array  of interaction codes.   

Barbieri has pointed out the importance of codes in living systems generally; he proposed that life emerges from codes that enable the maintenance and the development of structures and processes, including the genetic code and its expression; on this basis, he developed the concept of “codepoiesis”, the idea that living systems function to preserve organic codes and to evolve by developing new codes.  Barbieri defines a code as  “a mapping between the objects of two independent worlds” ~\cite{barbieri2015codeBiologyBook,barbieri2012codepoiesis}. Species interaction codes, in contrast, are not 
mappings between
``independent worlds''; rather they are descriptions of sets of mutually dependent interactions that link  organisms to their specific environments and ecosystems.

\begin{figure*}[t]
	\centering
	\includegraphics[width=0.9\linewidth]{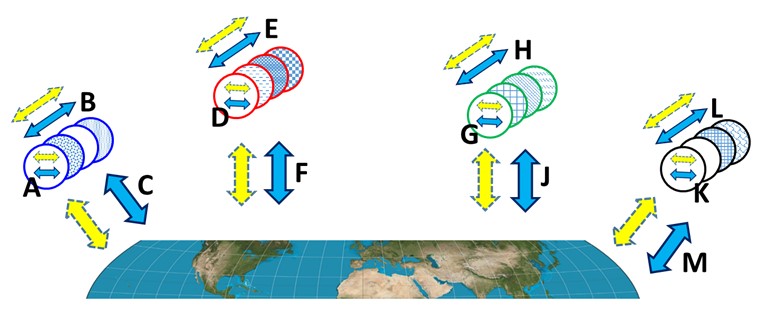}
	\caption{Species are defined by codes of core interactions. Organisms (shown as circles), are grouped in 
	species (distinguished by 
	blue, red, green and black borders), 
	where each species is defined by the set of its core sustaining essential interactions (shown as blue arrows); these interactions include those that are internal to each organism in the species, between members of the species, and with other species and with the inanimate environment. 
	 Thus, the three sets of interactions in the species interaction code of each species (internal, intra-species and external) are respectively marked with \{A, B, C\}, \{D, E, F\}, \{G, H, J\}, and \{K, L, M\}. 
	Species also engage in circumstantial interactions
	(shown as yellow dashed arrows), which are not part of the species interaction code.  Organisms within a species vary (shown as different fill patterns). 
}
	\label{fig:SIC}
\end{figure*}

Figure~\ref{fig:SIC} schematically summarizes the structure of the biosphere manifested through species codes of interactions. 

In principle, each species could be characterized by a particular book of interactions. 
A detailed list of a species interactions for even the ``simplest'' of species would challenge experts.  We suggest, however, that a pairwise perspective of species  might help clarify the concept of a species interaction code:
given two related species, we might focus only on 
detectable interactions that distinguish the pair.

\subsubsection*{Voles and Crows}
 
Here are two examples of related species that can be distinguished by a few differences in their interaction codes.
 
\textbf{Voles:} 
The species termed prairie vole (\textit{Microtus ochrogaster}) and meadow vole (\textit{M. pennsylvanicus})  look very much alike, but the two species differ markedly in reproductive and social behaviors: prairie vole males are largely
monogamous and social while meadow vole males are polygamous and solitary~\cite{gruder1985VoleSpecies}.  
These interaction patterns  are components of the codes that distinguish the two species.
But meadow voles can be induced to express prairie vole  interactions: 
Experimental insertion of a vasopressin receptor transgene into a specific site in the ventral forebrain of adult male meadow voles changes some of their reproductive and social behaviors to appear more similar to those of prairie voles; they become monogamous and friendly~\cite{young2004voleGeneVasopressin}.    
 
There is yet no information about the codes employed by female voles that distinguish the two different species. Presumably, 
the females of each species are 
attracted to interact with particular male behaviors.
Other interaction code differences are likely to account for interactions with the different environments of these species. 

\textbf{Crows:} 
Hooded crows (\textit{Corvus cornix}) and carrion crows (\textit{C. corone}) are very similar genetically (99.72 percent identical) to the point that they can produce fertile hybrid offspring~\cite{wolf2010crowSpecies}.  The two species are in contiguity in large areas of Europe; why has not one species dominated and eliminated the other? 
What factor maintains two closely related species in the same physical environments?  It turns out that the 0.28 percent genetic difference between the species includes the degree and pattern of feather pigmentation~\cite{poelstra2014genomicLandscapeCrows}; the two species, distinguished only by their appearance
(carrions are totally black; hoodeds are partly grey),
live in peace, defying Darwinian competition.  It seems that crows prefer to mate with partners who look like their parents~\cite{metzler2020assortativeMateChoice}.  Thus, 
the two species can be distinguished 
by a single interaction code determinant of what Darwin has termed sexual selection~\cite{darwin1860originOfSpecies}; what works, works.

\section{Reproduction and \\ Metabolism as Essential \\ Interactions} \label{sec:reproductionAndMetabolism}

Living systems manifest a great variety of interactions; 
however, 
the species interaction codes of all species include 
two essential properties:
their ability to reproduce their kind and their ability to metabolize the energy 
and building blocks
they require for maintenance and reproduction in their particular environment.
Quite simply, species whose constituent organisms 
are not collectively capable of the interactions that 
metabolize 
and reproduce the species cannot survive~\cite{dupre2013varietiesB}.

Obviously not every organism within a species 
need metabolize and reproduce: 
organisms may exist in states of suspended animation (deep hibernation, dry seeds, spores) and only certain organisms may engage in reproduction.  But metabolism and reproduction are interactions essential to the species as a collective whole,  even though different species may carry out these essential interactions in different ways. 
 
\section{The Role of Species}\label{sec:roleOfSpecies}
  
Life, like matter, must adhere to the physical laws of nature~\cite{cohenMarron2020survivalOfTheFitted}. Life, in its dependence on information and interaction, must accommodate
the second law of thermodynamics, which dictates that 
information---ordered structure---will deteriorate spontaneously into disorder.
One may argue that living systems are \textit{open systems} and so  
may be able to resist
the dictates of the second law;
nevertheless, all multi-cellular organisms die.  Boltzmann himself, and years later, Schroedinger, have called attention to the paradox
associated with the emergence of order and life 
~\cite[Chapter 3]{boltzmann2012SecondLaw},\cite[Chapter 6]{schrodinger1944WhatIslife}.
It is an observable fact that 
the persistence of life is accompanied by the re-production 
of its necessarily moribund organisms along with their metabolism.

How do 
 reproduction and metabolism persist
in a realm of universal individual death?
Clearly, non-reproducing singletons, do not last. Moreover, the loss of the singleton interrupts the networks in which the singleton acts.
The institution of species provides one answer---species feature functionally replaceable singletons.

A single organism becomes a multiplicity as it reproduces. 
And multiplicity helps deal with entropy; 
the reproducing collective obeys the dictates of entropy, but the collective whole replaces organisms lost to the species by death. 

From this perspective, we reason that the existence of multi-cellular life in a given environment requires functionally similar organisms in the aggregate framework of species. Individual organisms, by virtue of entropy, may come and go; only a collective species persists indefinitely in its environment---or at least until replaced by  evolution.

The fossil and genetic records support the conclusion that multi-cellular life has appeared in the framework of evolving species for hundreds of millions of years; although one might possibly imagine other ways 
that the biosphere could have evolved
to respond to the  inevitable death of individual organisms, no equivalent to the species framework of multi-cellularity has yet been detected. 

To paraphrase Darwin: we may say that multi-cellular life itself is the origin of species; if there be life, it must be in the form of a multitude of similar organisms organized as species. And, as we propose here, the members of a species are defined by their joint interaction code.  From this viewpoint, a species interaction code is essential to multi-cellular life.

We have discussed species, interaction codes and the processes of encoding and decoding.  We are now prepared to explore the possibility that evolution computes species using 
natural autoencoding.
First, we shall briefly describe artificial autoencoding by computer, and then we shall apply the autoencoding concept
to the natural autoencoding of species and evolution.

\section{Artificial Autoencoding}\label{sec:autoencoders}
 
Autoencoding is a term associated with artificial intelligence, machine learning, and artificial neural networks~\cite{kramer1991nonlinearAUTOENCODING}\cite{GoodfellowBengioCourville2016DeepLearningBook}.

An artificial
autoencoder is a computer program that extracts
the defining features of the individuals in a given population, and then represents each individual
as a set of values in a feature vector, 
or array. 
This \textit{code} and its formative \textit{encoding}, 
generated by artificial autoencoding, constitute a compact representation of the population and its individuals. 

A typical artificial autoencoder is a neural network that, through an interactive training process, 
establishes encoding and decoding computations and the associated code.
These computations can be used to encode each individual in a population, and to subsequently reconstruct each encoded individual from its respective code.

\begin{figure*}[t] 
	\centering
	\includegraphics[width=1.0\linewidth]{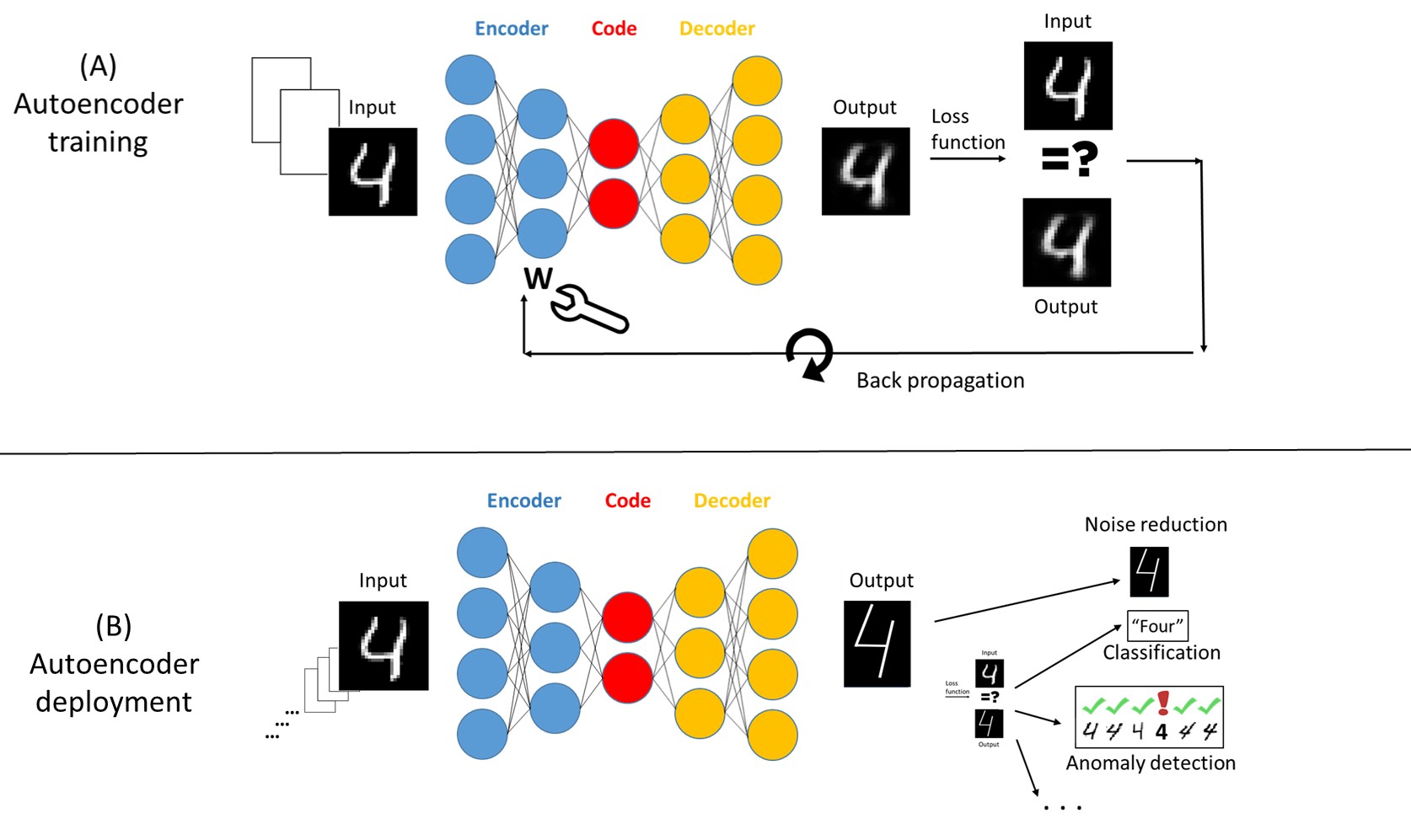}
	\caption{Artificial autoencoding. 
	 Typical autoencoders include three network-based elements: the encoder (blue circles), the code (red circles), and  the decoder (orange circles). Individual inputs (handwritten digits, for example) are fed into the encoder, encoded as values in the code feature vector, and then reconstructed by the decoder. \\
	 During training (A), the differences between the output and the input are computed by a loss function, and, in an optimization process, the weights W of the connecting lines in the autoencoder's neural net are adjusted to
	 minimize the reconstruction loss. 
	 This is repeated using a finite set of examples. The process is termed backpropagation, 
	 and is often done using a gradient descent method.
	 \\
	 ~~~Once training is completed, the autoencoder is deployed (B) to perform its application task. Encoding and decoding are now done using the fixed code and edge weights to process an unbounded number of inputs from the domain of interest. \\
	 The essence of artificial autoencoding is 
	 the creation and reshaping of the processes of encoders and decoders and thus of an entire working code system.}
	\label{fig:artificialAutoencoding}
	
\end{figure*}

The machinery of an artificial autoencoder includes four elements:
(1) \textit{The encoder} receives input data regarding selected individuals, 
such as pixels of an image, audio signals, or  measurements from some problem domain; 
the encoder outputs the learned 
feature vector with individual value assignments; this feature vector  constitutes (2) \textit{The code}.  (3) \textit{The decoder} accepts the code representing the encoding of a particular individual, and reconstructs the original input, such as the image or the sound segment. (4) The fourth element is the training algorithm that builds the encoder and the decoder. 
See Figure~\ref{fig:artificialAutoencoding}

In the process that builds the encoder and decoder, the  inputs and outputs are compared using a \emph{loss function} to determine how close the reconstructed outputs are to the respective inputs. The 
internal parameters of 
the encoder and the decoder, which are commonly built using neural networks,
are then adjusted and tuned, in an optimization process termed backpropagation, 
usually carried out by gradient descent, 
to minimize
the loss function.  

In artificial autoencoding, the training 
is unsupervised: the data are not labeled, so the autoencoder does not know what it is encoding. 
It is only required 
that the outputs 
be very similar to the corresponding inputs, specified by the loss function.

Once trained on a representative sample of a population, 
the autoencoder is able to encode and faithfully reconstruct many inputs from this population.
Furthermore, certain autoencoders, termed variational, can use the code to generate new entities to be included in that population~\cite{kingma2013VariationalAE}.

Artificial autoencoders enable many uses, including face recognition, 
image search, 
cleaning out image data by removing insignificant ``noise'', anomaly detection, classification and more.

The array of features that comprises the numerical code may or may not include  traits  that a human observer would 
intuitively use to compactly describe the individual. Hence this code vector is often referred to as the  \emph{latent}, invisible vector, where only a properly trained decoder can ``understand'' its features. 

Consequently, artificial autoencoding can function without its human operators assigning meaning to the details of the process, that is, how the multitude of  connection weights and non-linear functions relate to the problem at hand.  
Autoencoding takes place in a ``black box'', as it were,
in which 
humans choose the network architecture and the activation functions, select the input, and develop the loss function.  Autoencoder interpretability and explainability are still  areas of research.

The  opacity of autoencoding is important for our understanding of natural autoencoding of species and
biosphere evolution, described below.
Natural autoencoding takes place without any goals or processes selected by external agents. 

\section{Natural Autoencoding}\label{sec:autoencoding}

If codes define species, it would not be unreasonable to consider whether the formation and
evolution of species might involve some form of autoencoding. 
\textit{Natural autoencoding} would be an apt term if species evolution were to include 
the establishment of species code systems.

Let us analyze code, encoding and decoding in the contexts of life and evolution.

\subsubsection*{Natural codes}

We have defined a code as a description of interactions, and not as actual interactions. What then constitutes the species code book?  In what form might  a biological code reside?

In our definition of \textit{interactions}, we observed that sustained 
interactions are marked by \textit{repetition and sequence} (Section~\ref{sec:defs}). 
Accordingly, we propose that the molecular and behavioral features of the organism that enable the repetition of interactions constitute 
a description of such  interactions.
In other words, an infrastructure that anticipates 
a set of interactions is a code whose decoding materializes those interactions.
A reusable biological network is a biological code ``written'' for decoding. 

We propose that the species code book is  a composite of three forms of foundational information, 
both structural and dynamic: 
(1) the species germ-line genome; (2) the species physiology; and (3) the arrangement of the species within a given ecosystem.

The germ-line genome is a diversity of DNA sequence information distributed within the population of organisms composing the species.   
But the germ-line genetic code alone cannot serve as the code book of the species; genes alone are not a readable record;
genes have meaning only when  expressed~\cite{cohenAtlanEfroni2016geneticsAsExplanation}.

Danchin asserts that genetics alone is insufficient to account for biological function and that the underlying infrastructure of the whole cell, which he likens to a computer operating system, is essential~\cite{danchin2018CellOSPaleome20YearBacillusAnnotation}.

The ways genes get expressed, 
are not encoded directly 
in DNA sequences: a single gene TNF-$\alpha$,
for example, 
gets expressed differently in embryonic development, inflammation, healing, immune system reactions, cancer, aging, and other contexts~\cite{romanowska2021TNFrolesPregnancy}.  

TNF-$\alpha$ is not exceptional; most if not all proteins are pleiotropic---a single protein (derived from a single DNA sequence) will perform different functions in different contexts
~\cite{cohen2000tendingAdamsGarden}.  Moreover, different segments of a single gene sequence can be translated
differently to express different proteins~\cite{cohen2016updatingDarwin}.

Another example of extra-genetic information
is in cell division: the membrane of the daughter cell is built from the membrane of the dividing cell.
~\cite{mazia1961mitosisChapter}.

Consequently, the species code book must also include the molecular and physiological arrangements that describe the potential core interactions of the species---a description of \textit{species physiology}. 
Consider bee species: the queen bee and the worker bees carry genes involved in enzymatic interactions, 
but there are no genes that directly encode honey.
Honey, which is one of the outputs of interactions in the species interaction code of bees,
is produced by worker bees 
in contexts in which particular enzymes and metabolic pathways get activated repeatedly in sequence.

Organisms also require ecological arrangements along with their genomes and physiology to survive. Bees, for example, need flowers, certain weather profiles, and other information encoded within bee ecosystems. 
Thus the interaction with flowers which is part of the species interaction code of bees and of some flowers, 
is not directly coded in either the bees' DNA or the flowers' DNA, but is distributed. 

Bees are only one example, the code books of all species include the genetics, physiology and ecology of the species.

\subsubsection*{Natural encoding and decoding}

The species infrastructure codebook is written
by actual species interactions. 
Those interactions maintain the organisms of the species, strengthen the network infrastructure of the species and render it suitable for 
repetition---this constitutes
the species interaction code.  

The success of the input interactions confirms their potential  to continue to maintain the organisms of the species in future rounds of interaction.  In this way, input interactions are built into the species interaction code that forms a substrate for subsequent decoding.

Natural decoding is 
the continuing interactions of the organisms of the species in the species environment---genetic, 
physiological and ecological.

\subsubsection*{From ecosystems to the  biosphere}

Every species that survives does so thanks to its integration within an ecosystem that provides the essential matter and metabolic energy on which the organisms of the species depend for survival.

Most plant species survive by exploiting sunshine, 
soil
and water; insects may exploit plants and animals; animals may survive in predator-prey relationships; bacteria and fungi may exploit plants, animals, insects and other single-celled organisms. 

But beyond relatively local ecosystems, species ecosystems are also integrated into larger ecosystems that include multiple species in complex relationships with other species;
arrays of species link multiple levels of producers and predators 
and depend
on energy and other resources passing between and within the various species.

The ultimate output of the process is the decoding of the global array of species and environments that together constitute the biosphere ecosystem.
Thus the global
output of species decoding is the maintenance 
and the evolution
of the species and of the role of the species in the biosphere.  

\begin{figure*}[t] 
	\centering
	\includegraphics[width=1.00\linewidth]{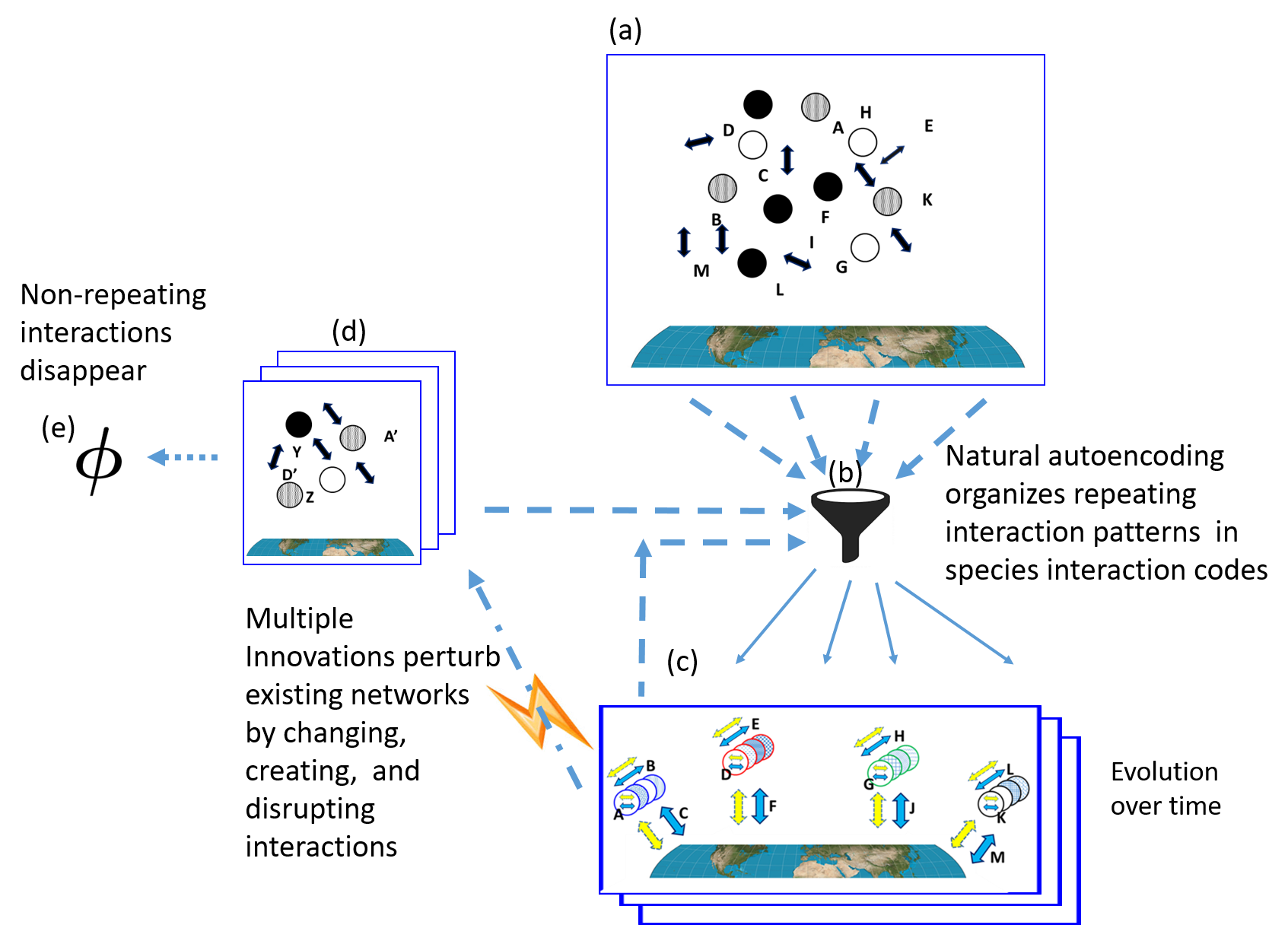}
	\vspace{0.5cm}
	\caption{Evolution by natural autoencoding. 
	The plethora of interactions taking place in the biosphere, the upper rectangle marked (a), are funneled (b) into networks of repeatable  species interaction codes, the rectangle marked (c).  Innovations, represented as a lightening bolt, can perturb or destroy species interactions, the rectangle marked (d) ; two outcomes can take place: new fitted, repeatable interactions can arise and be funneled into modified or new species; 
	unfitted, non-repeating interactions disappear (e). 
	} 
	
	\label{fig:naturalAutoencoding} 	
\end{figure*}

Figure~\ref{fig:naturalAutoencoding} summarizes evolution by natural autoencoding.   

\subsubsection*{Integrated encoding and decoding}

 Conceptually, we treated natural encoding and decoding as separate processes. 
 The interactions of living systems, however, are integrated into  composite networks; it would be difficult to label a given reproductive, developmental or metabolic process as purely encoding or purely decoding.  

For example, a sequence of DNA is decoded into a linear amino acid sequence, which itself is a coded description that encodes a functionally folded protein. 
This protein may then serve as an interaction code that is subsequently decoded into a structural protein, an enzyme or antibody that,
in turn,  
can help encode additional metabolic, immune or social interactions that maintain and protect the species organisms and the ecosystem.

In~\cite{cohenEfroni2019immuneStateComputation}, a machine learning process accounts for the encoding of the state of the body  by the mammalian immune system.
  
\begin{figure*}[t]
	\centering
	\includegraphics[width=0.95\linewidth]{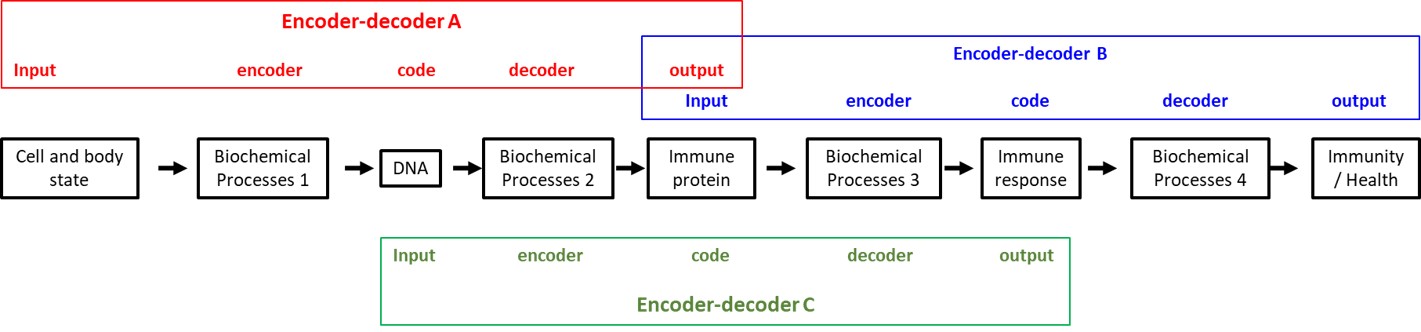}
		\vspace{0.5cm}
	\caption{Linked assemblies of biosphere encoders and decoders. 
    Multiple 
    encoder-decoder pathways,
    including processes and structural
	entities, operate in parallel and are intertwined in a variety of ways forming chains and networks.  
	Here for example,
    the immune system proteins are decoded from DNA by encoder-decoder A (shown in red above its constituent entities).
    Encoder-decoder B (blue) encodes the immune protein into an immune response pattern which is then decoded to generate 
    health. 
    Encoder-decoder B may be connected to additional entities and processes (not shown). 
    Encoder-decoder C (shown in green below its constituent entities) can be seen as encoding a DNA sequence into a protein molecule code, which is then decoded into an immune response.
    More generally, every interaction, input, code, or output of a given natural encoder-decoder pathway
        may also serve a function in another encoder-decoder pathway.
    The inputs to each encoding and decoding process may include additional entities beyond the ones shown here.
    Furthermore, each one of the many steps that constitute each encoder or decoder is an interaction in its own right,
    with its own description of input, encoder process, code, decoder process and output.}
    	\label{fig:overlappingEncoderDecoder}
\end{figure*}

More generally, every interaction, input, code, or output of a given natural 
encoding-decoding 
process
may also serve a function in another 
natural encoding-decoding process
(See Figure~\ref{fig:overlappingEncoderDecoder}).  
Every encoding or decoding process is itself an interaction (or set of interactions) associated with its own code.  

Our concept of natural autoencoding 
is based on the realization that the biosphere is sustained 
by repeated interactions.
Continuous encoding and decoding are central in maintaining and evolving this repetition. 
Encoding and decoding in the context of the biosphere, in contrast to the computer, 
refers to the indefinite repetition of biological interactions on planet earth in which one set of reactions---input information---is represented by a code that 
is decoded 
simultaneously 
into another set of reactions---output information.   
\subsubsection*{Constraints}
Interactions in general are organized by limitations, \textit{constraints}, imposed on the  interactors and on the environment; no stable structures can emerge when degrees of freedom are not limited~\cite{grotzinger1995ConstraintsOnEvolution}.
Moreover, constraints decrease entropy by limiting the state-space of the system.
Interactions result from constraints that channel the interactors to meet and interact.  
In fact,
interactions themselves generate new constraints on what may follow. The cell membrane, for example, is formed by interactions between lipids, proteins and other molecules that effectively establish the boundary of the cell. This boundary keeps the contents of the cell together; localization is required for any interaction.

The force of gravity is a constraint that localizes land, water and atmosphere as media for life.

\subsubsection*{Housekeeping autoencoding}

We distinguish \textit{housekeeping autoencoding} from \textit{evolutionary autoencoding}. 

Housekeeping autoencoding refers to 
existing species interactions that have not been perturbed by innovations that change species codes.
Biological ``business as usual'' is housekeeping---maintaining the house in a state of homeostasis.

Natural housekeeping autoencoding is similar to the error correction and noise reduction applications of artificial autoencoding 
in that the decoding process restores the integrity of the input. 
In artificial autoencoding noise reduction is computed in a trained autoencoder using the learned
weights of the neural net (see Figure~\ref{fig:artificialAutoencoding}). 
In natural housekeeping autoencoding,
the output is generated by subjecting the input to programs for DNA repair~\cite{sancar2004DNARepair}, 
cancer suppressor mechanisms~\cite{soussi2000p53},
immune reactions, reproduction and programmed death~\cite{cohen2000tendingAdamsGarden}, among others.  These interaction programs are included in existing species
interaction codes. Housekeeping autoencoding maintains the 
state of the individual by preserving and restoring a 
healthy, sustainable 
body state.

But preservation and restoration
of individual homeostasis are not evolution. 
Evolutionary autoencoding refers to the evolution of new species interaction codes.

\subsubsection*{Innovations and evolutionary  autoencoding}

Changes in species interaction codes emerge from innovations that lead to novel interactions in encoding and decoding. 
Innovations are perturbations that are not accommodated within the code of interactions of the existing species. 

Perturbations and variations can take place both in the present code and in the environment---for example, genetically variant members of a hominoid species migrated out of Africa to evolve into the Neanderthal species in the European environment~\cite{mellars2004neanderthals}. 

An innovation can enter the biosphere in a variety of ways and forms, be it a molecular mutation, an infecting pathogen, an invading species, a cancer cell, a change in nutrients or in solar radiation, a natural cataclysm or a social or technological invention; witness, for example, the industrial revolution and global warming~\cite{rosenzweig2008anthropogenicChangesEnv}.

If an innovation is not integrated into a fitted configuration 
within  networked species interaction codes, 
an unfitted interaction state can emerge; unfittedness
can negatively affect molecules, cells, organisms, species, and ecosystems; 
an innovation that does not integrate into a repeatable network
will ultimately fail to survive and disappear. See Figure~\ref{fig:naturalAutoencoding}.
\section{Sexual Reproduction, \\Entropy and Natural\\ Autoencoding}\label{sec:sexrep}

\subsubsection*{The challenge of sexual reproduction}

Sexual reproduction has long presented a  problem for the neo-Darwinian theory of evolution  ~\cite[Ch.3]{dawkins1976selfishGene}, \cite[p.265]{ridley1993Evolution}, \cite{livnatPapaDimitriou2016SexAsAnAlgorithm}. Bell termed sexual reproduction the queen of problems in evolutionary biology~\cite[Ch.1]{bell2019masterpieceOfNatureEvolutionAndGeneticsOfSexuality}.

To summarize the problem: natural selection teaches that evolution acts as an optimization process of individual fitness; 
yet, no matter how fit an individual may be, sexual reproduction, which involves random genetic recombination of parental genes, guarantees that one’s offspring will never inherit one’s exact genomic fitness. It seems counterproductive to select fit individuals and then to randomly disperse their genomes in the next generation. 
Despite many hypotheses the problem is still open.

\subsubsection*{Autoencoding and sexual reproduction}

The natural autoencoding mechanism of evolution, in contrast to Darwinian natural selection,
is not thwarted by the genetic randomization inherent in sexual reproduction;  
on the contrary, sexual reproduction, we reason, is essential in the natural autoencoding of most multicellular organisms. 

As we wrote above, the survival of a species depends on the replacement of dead organisms by the reproduction of still living organisms (Section 4). The creation of a newborn results in
a significant increase in order and complexity;  this decrease in entropy is compensated by the increase in entropy generated by  
 the unpredictable, random genomic recombination of the germ cells of the two parent organisms. Sexual reproduction fulfills a law of nature. 

\subsubsection*{Diversification, natural autoencoding, and sex}
The second law of thermodynamics dictates that organized structures will in time deteriorate and diversify.  Diversification dismantles the optimum, the goal of natural selection.  However, 
diversification is actually a necessary factor in natural autoencoding.  
An examination of artificial machine learning can help explain why.

Most forms of computer machine learning algorithms rely on inputs featuring
randomly selected, diverse manifestations of 
the element to be learned: 
for example, artificial autoencoding of images of dogs begins by feeding the algorithm with many diverse representations of dogs~\cite{GoodfellowBengioCourville2016DeepLearningBook}; 
teaching a computer to distinguish dogs from other entities (a classification task)  may also require 
feeding diverse images of entities that are not dogs. 
A single dog photo, or even a million photos of a single dog will not suffice.
The computer program needs to extract  from many diverse
photos of dogs and other entities
the core features that characterize ``dogness''. 
If the diversity of  available input data is insufficient, some learning algorithms 
perturb or add random noise to the original data;
this challenges the learning process to identify relevant features. 

Natural autoencoding of a    
species interaction code, 
like computer machine learning, 
requires experience with randomly diverse examples of genomes
and phenotypes borne by members of the species that, despite their genetic differences, 
thrive in the species environment. 
Sexual reproduction enables the continuous input into the species environment of organisms bearing workable arrays of genomic diversity. 

Moreover, random diversification by sexual reproduction defines the functional extent of genome variation operating in the species. Failure of sexual interactions to generate reproducing offspring limits the effects of the second law of thermodynamics which guarantees that random genetic mutations will occur. 
Sexual reproduction culls the species of ``bad'' genes and gene combinations;  genomes of sexually
reproducing organisms that fail to be propagated into the next generation are weeded out in the  process of reproduction.

Sexual reproduction not only establishes the functional diversity of a species' DNA genome; the sex act 
also tests physiological, social and ecological codes within the species:
Attraction, courting, nesting, 
and rituals, 
physical and symbolic,   
mark the sexual reproduction of many species.
Sex establishes, maintains and tests many interaction codes in the given species' environment.

Sexual reproduction may be 
a problem for evolutionary concepts based on individual optimization, but not for a concept  
of evolution based on
natural autoencoding of ``what works''. 
Sexual reproduction removes species interactions that 
do not 
work in the species environment and it does so by 
design, and not by unpredictable accident.
Sexual reproduction thus  enables the species as a whole to autoencode itself genetically, physiologically and ecologically.

We propose that 
sexual reproduction is essentially universal in multi-cellular species 
because the repetition of living interactions must continue despite the inevitable diversification dictated by the second law of thermodynamics.   
Sex, from this perspective is foundational, not paradoxical. 

As we mentioned in Section~\ref{sec:speciesInteractionCode}, the concept of bacterial species is controversial~\cite{chun2018prokaryoteSpecies}. Moreover, bacteria do not engage in sexual reproduction; however, it has been suggested that  horizontal gene transfer may play a
 role in defining the borders of species in bacteria
~\cite{delaCruz2000horizontalGeneTransferProkaryotesSpecies}, along with compensatory genomic diversification and increased entropy.

\subsubsection*{Summary of sexual reproduction, entropy and natural autoencoding}

\begin{figure*}[h] 
	\centering
	\includegraphics[width=0.9\linewidth]{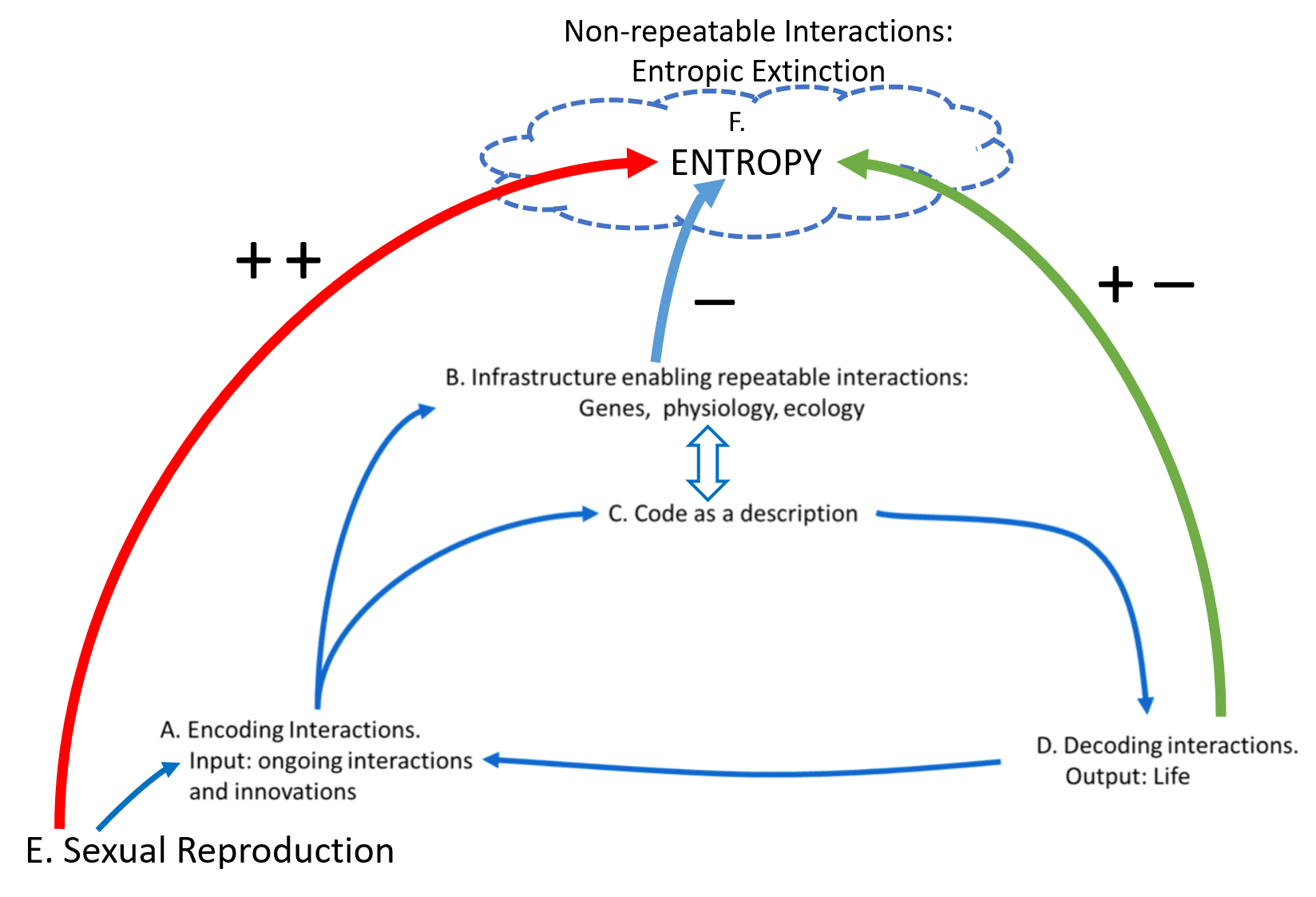}
	\caption{Schematic view of natural autoencoding, sexual reproduction and the relevant effects on entropy. See text for details.}
	\label{fig:schemaAEwithEntropy} 	
\end{figure*}

Figure~\ref{fig:schemaAEwithEntropy} summarizes the links 
between natural autoencoding, sexual reproduction and entropy.
The mechanism of natural autoencoding is outlined in the figure in the items marked A, B, C and D; the role of sexual reproduction is labeled E, and entropy is marked F; 
the arrows designate influences and relationships.
Encoding interactions, marked A in the figure, 
are composed of both ongoing  interactions and innovations, such as molecular mutations, new invading parasites or physical perturbations and environmental changes.

This input proceeds
in two separate pathways: 
A to B and A to C. 
Pathway A to B 
marks those repeatable, surviving genetic, physiological and ecological interactions that compose the network infrastructure that enables the repetition of interactions; this infrastructure,
constitutes a description of species interaction codes (shown as a wide bidirectional arrow connecting B and C). 
Innovations that integrate into supporting networks may change species interaction codes and thus generate new species. 
Encoding interactions not only generate the species interaction codes, input  interactions also activate existing infrastructure networks, shown as the connecting arrows from A to C and C to D.  
This decoding of the species interaction code infrastructure realizes the outputs that constitute
life. The interactions of life (D) feed back 
into the input  (A) that generates and activates the species interaction codes (C) of the living, evolving biosphere (A-D).

The organized biosphere (A-D), like all of material existence,  constantly generates a degree of compensatory disorganization or lost energy.

The enabling structure of life and its natural autoencoding (A-D) reduces,
entropy; this is shown as the thick blue arrow from B to F,
labeled with a minus sign ($-$).
sexual reproduction (E), 
which is an input into A, compensates life’s overall decrease of entropy by enhancing the entropy of 
organismal replacement (F) 
through random genomic diversification, marked by the 
thick red arrow ($++$).  

The interactions of life that organize the biosphere also contribute an increase in entropy by lost heat, degradation, illness, mortality, individual differences and destruction of organization (thick green arrow marked $+$~$-$).  
Non-repeatable interactions are lost to entropic extinction.

\section{Comparing Artificial and Natural Autoencoding}\label{sec:differences}
\begin{figure*}[t] 
	\centering
	\includegraphics[width=1.0\linewidth]{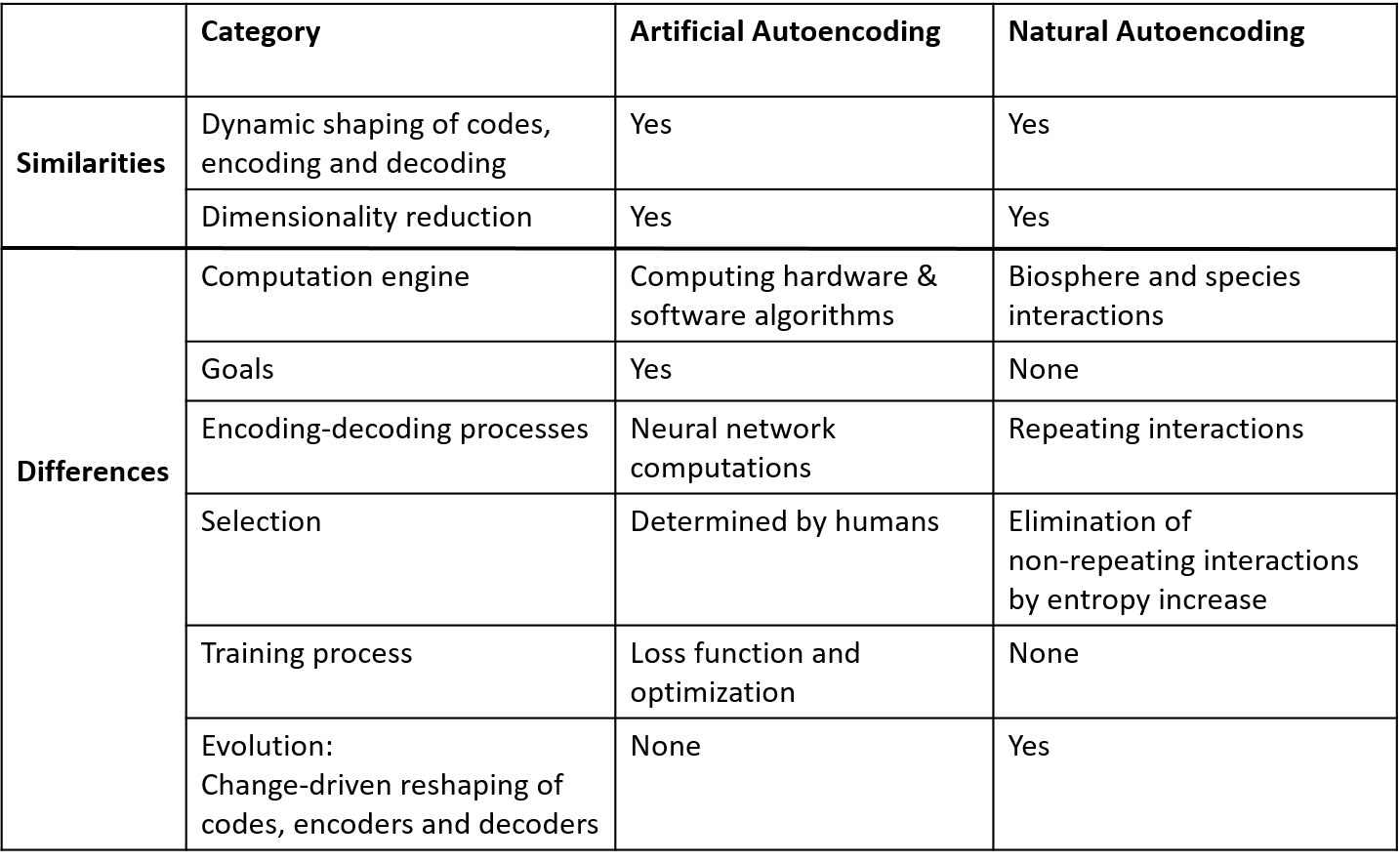}
	\caption{Summary comparison of natural and artificial autoencoding. See text for details.}
	\label{fig:ComparingArtificialNaturalAE} 	
\end{figure*}

Figure~\ref{fig:ComparingArtificialNaturalAE} summarizes the similarities and differences between artificial and natural autoencoding. Both processes shape code systems and both reduce the dimensionality of input data to essentials. However, natural autoencoding aims at no designated goals and has no training process involving loss functions and optimizations. Natural autoencoding does not use computing hardware or software, 
but is the outcome of biosphere and species interactions in which repeating interactions
become species interaction codes; non-repeating interactions are eliminated selectively by entropy.  Moreover, natural autoencoding, in response to innovations, spontaneously evolves its encoding, codes or decoding processes and interactions.  

Evolution seems to have experienced natural autoencoding millions of years before humans developed artificial autoencoding.
We arrived at
the concept of natural autoencoding by way of  artificial computer autoencoding.
However, from the perspective of evolutionary time, artificial autoencoding, is a ``non-conventional'' variant of the natural process that preceded it.  

\section{On Evolution as a Machine Learning Process}

Recently, Vanchurin and colleagues 
have applied machine learning concepts and thermodynamic principles to develop a theory of evolution as multilevel learning~\cite{vanchurinKoonin2022EvolutionAsMultilevelLearning,vanchurinKoonin2022thermodynamicsOfEvolution}.  

The main differences between the ideas of natural autoencoding and the theory of Vanchurin and colleagues are as follows:  

Vanchurin and colleagues state that the biosphere learns to compute a fitness function, uses it to compute the loss function, and then optimizes this loss function, for example: 
\textit{``We make the case that loss function, which is central to the
learning theory, can be usefully and generally employed as the
equivalent of the fitness function in the context of evolution.''}~\cite{vanchurinKoonin2022EvolutionAsMultilevelLearning}. In contrast, the  evolutionary mechanism of autoencoding proposed here
does not compute fitness, and does not use a loss function. 
Furthermore, in the computations carried out by natural autoencoding in response to innovations
there is no optimization: ``what works, works''.
Natural autoencoding is based only on the
observed preservation of repeating interactions.

\section{Humans and the Biosphere}\label{sec:humansAndBiosphere}

The rapid expansion of the human population in the past ten thousand years owes its onset to the domestication of species of plants and animals by humans; 
humans chose to propagate only those species innovations that satisfied  perceived human needs.
 Whether or not one accepts Darwin's idea of Natural Selection as a ``law of nature''~\cite{bradley2022LawOfNatureCauseEffect,reed1981lawfulnessOfNaturalSelection,byerly1983naturalSelectionAsALawPrinciplesandProcesses}, the ``natural right'' of domination by the ``fittest'' has influenced many aspects of human culture including ethics, economics, governance, racial relations, 
 social organization, and education \cite{mayr2000darwinsInfluenceOnModernThought,
 wyllie1959socialDarwinismAndTheBusinessMan,auerswald2003DarwinianSeas,browning2017fittestInEducation,bergman2014darwinNazismEugenicsRacismCommunismCapitalismSexism}.
The centrality of domination 
in natural selection is problematic both for our understanding of the biosphere and for our behavior within it.

Thomas Kuhn has pointed out the blinding power of entrenched paradigms in science~\cite{kuhn1970scientificRevolutions}. Traditional studies
of evolution assume survival-of-the-fittest as a given, even when they attempt to account for group cooperation 
~\cite{nowak2006fiveRulesEvolutionOfCooperation}.

Misunderstanding alone is tolerable; misguided action is not.
The spirit of domination 
underlies much of the irresponsible human behavior that is now 
changing the biosphere.  

Natural selection is not sufficiently sensitive to the worldwide
web of cooperative interactions among species and environments required to maintain
a biosphere friendly to the well-being of the human species.
We hope that an appreciation of 
Natural
Autoencoding and Survival-of-the-Fitted
will help 
support the movement to change
the  human interactions presently damaging the biosphere.

\section{Modeling and Simulation}\label{sec:future}
Here, we have introduced the concept of natural autoencoding; 
however reasonable, concepts alone do not suffice.  Further work needs to be done to support or refute the idea. 
We are now developing architectures and algorithms for computer modeling and mathematical definitions of natural autoencoding. 
We use standard and new artificial autoencoding techniques, including neural networks, agent based modeling, principal component analysis (PCA) algorithms, and property-preserving mathematical transformations of large data structures.
We do not claim that the forces and interactions of nature can be mapped directly to elements of any specific modeling technique. 
Nevertheless, these models can extend our understanding of the biosphere
and might even provide new tools in computer science. 

\section*{Acknowledgments}
We thank Yuval Bayer, Gil Egozi, Guy Frankel, David Harel, Eugene Koonin, Antonio López-Pinto, Yair Neuman, Eugene Rosenberg, and Smadar Szekely for valuable discussions and suggestions. 
This work was partially supported by research grants to David Harel:   Grant 3698/21 from the ISF-NSFC joint to the Israel Science Foundation and the National Science Foundation of China, and a grant from the Minerva foundation.

\section*{Author contributions}
IRC and AM contributed equally to the paper.

\section*{Competing interests}
The authors have no competing interests. 


\begin{thebibliography}{10}

\bibitem{auerswald2003DarwinianSeas}
P.~E. Auerswald and L.~M. Branscomb.
\newblock Valleys of death and {Darwinian} seas: Financing the invention to
  innovation transition in the united states.
\newblock {\em The Journal of Technology Transfer}, 28(3):227--239, 2003.

\bibitem{barbieri2012codepoiesis}
M.~Barbieri.
\newblock Codepoiesis---{The} deep logic of life.
\newblock {\em Biosemiotics}, 5(3):297--299, 2012.

\bibitem{barbieri2015codeBiologyBook}
M.~Barbieri.
\newblock {\em Code biology: A new science of life}.
\newblock Springer, 2015.

\bibitem{bell2019masterpieceOfNatureEvolutionAndGeneticsOfSexuality}
G.~Bell.
\newblock {\em The masterpiece of nature: {The} evolution and genetics of
  sexuality}.
\newblock Routledge, 2019.
\newblock Originally published in in 1982.

\bibitem{bergman2014darwinNazismEugenicsRacismCommunismCapitalismSexism}
J.~Bergman.
\newblock {\em The Darwin Effect: It's influence on Nazism, Eugenics, Racism,
  Communism, Capitalism \& Sexism}.
\newblock New Leaf Publishing Group, 2014.

\bibitem{blaser2014microbiomeRevolution}
M.~J. Blaser.
\newblock The microbiome revolution.
\newblock {\em The Journal of Clinical Investigation}, 124(10):4162--4165,
  2014.

\bibitem{boltzmann2012SecondLaw}
L.~Boltzmann.
\newblock The second law of thermodynamics.
\newblock In {\em {Theoretical Physics and Philosophical Problems: Selected
  writings}}. Springer Science \& Business Media, 2012.

\bibitem{danchin2018CellOSPaleome20YearBacillusAnnotation}
R.~Borriss, A.~Danchin, C.~R. Harwood, C.~M{\'e}digue, E.~P. Rocha,
  A.~Sekowska, and D.~Vallenet.
\newblock Bacillus subtilis, the model gram-positive bacterium: 20 years of
  annotation refinement.
\newblock {\em Microbial biotechnology}, 11(1):3--17, 2018.

\bibitem{bradley2022LawOfNatureCauseEffect}
B.~Bradley.
\newblock Natural selection according to {Darwin}: cause or effect?
\newblock {\em History and Philosophy of the Life Sciences}, 44(2):1--26, 2022.

\bibitem{browning2017fittestInEducation}
L.~Browning, K.~Thompson, and D.~Dawson.
\newblock From early career researcher to research leader: survival of the
  fittest?
\newblock {\em Journal of Higher Education Policy and Management},
  39(4):361--377, 2017.

\bibitem{byerly1983naturalSelectionAsALawPrinciplesandProcesses}
H.~C. Byerly.
\newblock Natural selection as a law: Principles and processes.
\newblock {\em The American Naturalist}, 121(5):739--745, 1983.

\bibitem{chun2018prokaryoteSpecies}
J.~Chun, A.~Oren, A.~Ventosa, H.~Christensen, D.~R. Arahal, M.~S. da~Costa,
  A.~P. Rooney, H.~Yi, X.-W. Xu, S.~De~Meyer, et~al.
\newblock Proposed minimal standards for the use of genome data for the
  taxonomy of prokaryotes.
\newblock {\em International journal of systematic and evolutionary
  microbiology}, 68(1):461--466, 2018.

\bibitem{cohen2000tendingAdamsGarden}
I.~R. Cohen.
\newblock {\em Tending Adam's Garden: evolving the cognitive immune self}.
\newblock Elsevier, 2000.

\bibitem{cohen2006informationallandscapesInArt}
I.~R. Cohen.
\newblock Informational landscapes in art, science, and evolution.
\newblock {\em Bulletin of Mathematical Biology}, 68(5):1213--1229, 2006.

\bibitem{cohen2016updatingDarwin}
I.~R. Cohen.
\newblock Updating {Darwin}: information and entropy drive the evolution of
  life.
\newblock {\em F1000Research}, 5, 2016.

\bibitem{cohenAtlanEfroni2016geneticsAsExplanation}
I.~R. Cohen, H.~Atlan, and S.~Efroni.
\newblock Genetics as explanation: Limits to the human genome project.
\newblock {\em Encyclopedia of Life Sciences.John Wiley \& Sons, Ltd:
  Chichester}, 2016.

\bibitem{cohenEfroni2019immuneStateComputation}
I.~R. Cohen and S.~Efroni.
\newblock The immune system computes the state of the body: crowd wisdom,
  machine learning, and immune cell reference repertoires help manage
  inflammation.
\newblock {\em Frontiers in immunology}, 10:10, 2019.

\bibitem{cohenMarron2020survivalOfTheFitted}
I.~R. Cohen and A.~Marron.
\newblock The evolution of universal adaptations of life is driven by universal
  properties of matter: energy, entropy, and interaction.
\newblock {\em F1000Research}, 9, 2020.

\bibitem{darwin1860originOfSpecies}
C.~Darwin.
\newblock {\em On the Origin of Species by Means of Natural Selection, or the
  Preservation of Favoured Races in the Struggle for Life}.
\newblock John Murray, 1860.

\bibitem{dawkins1976selfishGene}
R.~Dawkins.
\newblock {\em {The Selfish Gene}}.
\newblock Oxford university press, 1976.

\bibitem{delaCruz2000horizontalGeneTransferProkaryotesSpecies}
F.~De~la Cruz and J.~Davies.
\newblock Horizontal gene transfer and the origin of species: lessons from
  bacteria.
\newblock {\em Trends in microbiology}, 8(3):128--133, 2000.

\bibitem{doige2012typologyOfHeatAcrossTextbooks}
C.~A. Doige and T.~Day.
\newblock A typology of undergraduate textbook definitions of heat across
  science disciplines.
\newblock {\em International Journal of Science Education}, 34(5):677--700,
  2012.

\bibitem{dupre2013varietiesB}
J.~Dupr{\'e} and M.~A. O’Malley.
\newblock Varieties of living things: life at the intersection of lineage and
  metabolism.
\newblock In {\em Vitalism and the Scientific Image in Post-Enlightenment Life
  Science, 1800-2010}, pages 311--343. Springer, 2013.

\bibitem{gleick1993geniusFeynman}
J.~Gleick.
\newblock {\em Genius: The life and science of Richard Feynman}.
\newblock Vintage, 1993.

\bibitem{GoodfellowBengioCourville2016DeepLearningBook}
I.~Goodfellow, Y.~Bengio, and A.~Courville.
\newblock {\em Deep Learning}.
\newblock MIT Press, 2016.
\newblock \url{http://www.deeplearningbook.org}. Chapter 14.

\bibitem{grotzinger1995ConstraintsOnEvolution}
J.~P. Grotzinger, S.~A. Bowring, B.~Z. Saylor, and A.~J. Kaufman.
\newblock Biostratigraphic and geochronologic constraints on early animal
  evolution.
\newblock {\em Science}, 270(5236):598--604, 1995.

\bibitem{gruder1985VoleSpecies}
S.~Gruder-Adams and L.~L. Getz.
\newblock Comparison of the mating system and paternal behavior in microtus
  ochrogaster and m. pennsylvanicus.
\newblock {\em Journal of Mammalogy}, 66(1):165--167, 1985.

\bibitem{igamberdiev2018Leibniz}
A.~U. Igamberdiev.
\newblock Time and life in the relational universe: Prolegomena to an integral
  paradigm of natural philosophy.
\newblock {\em Philosophies}, 3(4):30, 2018.

\bibitem{judgeDodd2020Metabolism}
A.~Judge and M.~Dodd.
\newblock {Metabolism}.
\newblock {\em Essays in Biochemistry}, 64(4):607--647, 08 2020.

\bibitem{kingma2013VariationalAE}
D.~P. Kingma and M.~Welling.
\newblock Auto-encoding variational bayes.
\newblock {\em arXiv preprint arXiv:1312.6114}, 2013.

\bibitem{koonin2011logicOfChance}
E.~V. Koonin.
\newblock {\em {The Logic of Chance: the Nature and Origin of Biological
  Evolution}}.
\newblock FT press, 2011.

\bibitem{kramer1991nonlinearAUTOENCODING}
M.~A. Kramer.
\newblock Nonlinear principal component analysis using autoassociative neural
  networks.
\newblock {\em AIChE journal}, 37(2):233--243, 1991.

\bibitem{kuhn1970scientificRevolutions}
T.~S. Kuhn.
\newblock {\em The structure of scientific revolutions}, volume 111.
\newblock Chicago University of Chicago Press, 1970.

\bibitem{routledge2009metaphysics}
R.~Le~Poidevin, S.~Peter, M.~Andrew, and R.~P. Cameron.
\newblock {\em The Routledge companion to metaphysics}.
\newblock Routledge, 2009.

\bibitem{young2004voleGeneVasopressin}
M.~M. Lim, Z.~Wang, D.~E. Olaz{\'a}bal, X.~Ren, E.~F. Terwilliger, and L.~J.
  Young.
\newblock Enhanced partner preference in a promiscuous species by manipulating
  the expression of a single gene.
\newblock {\em Nature}, 429(6993):754--757, 2004.

\bibitem{livnatPapaDimitriou2016SexAsAnAlgorithm}
A.~Livnat and C.~H. Papadimitriou.
\newblock Sex as an algorithm: the theory of evolution under the lens of
  computation.
\newblock {\em Commun. ACM}, 59(11):84--93, 2016.

\bibitem{mallet1995speciesDef}
J.~Mallet.
\newblock A species definition for the modern synthesis.
\newblock {\em Trends in Ecology \& Evolution}, 10(7):294--299, 1995.

\bibitem{mayr2000darwinsInfluenceOnModernThought}
E.~Mayr.
\newblock Darwin’s influence on modern thought.
\newblock {\em Scientific American}, 283(1):78--83, 2000.

\bibitem{mazia1961mitosisChapter}
D.~Mazia.
\newblock Mitosis and the physiology of cell division.
\newblock In {\em The cell}, pages 77--412. Elsevier, 1961.

\bibitem{mellars2004neanderthals}
P.~Mellars.
\newblock Neanderthals and the modern human colonization of {Europe}.
\newblock {\em Nature}, 432(7016):461--465, 2004.

\bibitem{metzler2020assortativeMateChoice}
D.~Metzler, U.~Knief, J.~V. Pe{\~n}alba, and J.~B. Wolf.
\newblock Assortative mate choice and epistatic mating-trait architecture
  induces complex hybrid-zone movement.
\newblock {\em bioRxiv}, 2020.

\bibitem{nowak2006fiveRulesEvolutionOfCooperation}
M.~A. Nowak.
\newblock Five rules for the evolution of cooperation.
\newblock {\em Science}, 314(5805):1560--1563, 2006.

\bibitem{OED2022}
{Oxford English Dictionary (OED)}.
\newblock {Online Dictionary}.
\newblock 2022.
\newblock {https://www.oed.com} Definition of code.

\bibitem{poelstra2014genomicLandscapeCrows}
J.~W. Poelstra, N.~Vijay, C.~M. Bossu, H.~Lantz, B.~Ryll, I.~M{\"u}ller,
  V.~Baglione, P.~Unneberg, M.~Wikelski, M.~G. Grabherr, et~al.
\newblock The genomic landscape underlying phenotypic integrity in the face of
  gene flow in crows.
\newblock {\em Science}, 344(6190):1410--1414, 2014.

\bibitem{reed1981lawfulnessOfNaturalSelection}
E.~S. Reed.
\newblock The lawfulness of natural selection.
\newblock {\em The American Naturalist}, 118(1):61--71, 1981.

\bibitem{ridley1993Evolution}
M.~Ridley.
\newblock {\em Evolution}.
\newblock Blackwell Scientific, 1993.

\bibitem{romanowska2021TNFrolesPregnancy}
K.~Romanowska-Pr{\'o}chnicka, A.~Felis-Giemza, M.~Olesi{\'n}ska,
  P.~Wojdasiewicz, A.~Paradowska-Gorycka, and D.~Szukiewicz.
\newblock The role of {TNF}-$\alpha$ and anti-{TNF}-$\alpha$ agents during
  preconception, pregnancy, and breastfeeding.
\newblock {\em International Journal of Molecular Sciences}, 22(6):2922, 2021.

\bibitem{rosenberg2007coralSymbiosis}
E.~Rosenberg, O.~Koren, L.~Reshef, R.~Efrony, and I.~Zilber-Rosenberg.
\newblock The role of microorganisms in coral health, disease and evolution.
\newblock {\em Nature Reviews Microbiology}, 5(5):355--362, 2007.

\bibitem{rosenzweig2008anthropogenicChangesEnv}
C.~Rosenzweig, D.~Karoly, M.~Vicarelli, P.~Neofotis, Q.~Wu, G.~Casassa,
  A.~Menzel, T.~L. Root, N.~Estrella, B.~Seguin, et~al.
\newblock Attributing physical and biological impacts to anthropogenic climate
  change.
\newblock {\em Nature}, 453(7193):353--357, 2008.

\bibitem{rovelli2018realityInteraction}
C.~Rovelli.
\newblock {Carlo Rovelli — All Reality Is Interaction}, 2017.
\newblock \url{https://www.wnyc.org/story/59a21fbf4616dd86cb8fc341}; WNYC Radio
  Program; Accessed January 2023.

\bibitem{sachs2004evolutionOfCoop}
J.~L. Sachs, U.~G. Mueller, T.~P. Wilcox, and J.~J. Bull.
\newblock The evolution of cooperation.
\newblock {\em The Quarterly review of biology}, 79(2):135--160, 2004.

\bibitem{sancar2004DNARepair}
A.~Sancar, L.~A. Lindsey-Boltz, K.~{\"U}nsal-Ka{\c{c}}maz, and S.~Linn.
\newblock {Molecular mechanisms of mammalian DNA repair and the DNA damage
  checkpoints}.
\newblock {\em Annual review of biochemistry}, 73(1):39--85, 2004.

\bibitem{schrodinger1944WhatIslife}
E.~Schr{\"o}dinger.
\newblock {\em {What is life? The physical aspect of the living cell and
  mind}}.
\newblock Cambridge University Press Cambridge, 1944.

\bibitem{shannon1948informationTheory}
C.~E. Shannon.
\newblock A mathematical theory of communication.
\newblock {\em The Bell system technical journal}, 27(3):379--423, 1948.

\bibitem{simard2018mycorrhizal}
S.~W. Simard.
\newblock Mycorrhizal networks facilitate tree communication, learning, and
  memory.
\newblock In {\em Memory and learning in plants}, pages 191--213. Springer,
  2018.

\bibitem{soussi2000p53}
T.~Soussi.
\newblock The p53 tumor suppressor gene: from molecular biology to clinical
  investigation.
\newblock {\em Annals of the New York Academy of Sciences}, 910(1):121--139,
  2000.

\bibitem{vanchurinKoonin2022EvolutionAsMultilevelLearning}
V.~Vanchurin, Y.~I. Wolf, M.~I. Katsnelson, and E.~V. Koonin.
\newblock Toward a theory of evolution as multilevel learning.
\newblock {\em Proceedings of the National Academy of Sciences},
  119(6):e2120037119, 2022.

\bibitem{vanchurinKoonin2022thermodynamicsOfEvolution}
V.~Vanchurin, Y.~I. Wolf, E.~V. Koonin, and M.~I. Katsnelson.
\newblock Thermodynamics of evolution and the origin of life.
\newblock {\em Proceedings of the National Academy of Sciences},
  119(6):e2120042119, 2022.

\bibitem{wolf2010crowSpecies}
J.~B. Wolf, T.~Bayer, B.~Haubold, M.~Schilhabel, P.~Rosenstiel, and D.~Tautz.
\newblock Nucleotide divergence vs. gene expression differentiation:
  comparative transcriptome sequencing in natural isolates from the carrion
  crow and its hybrid zone with the hooded crow.
\newblock {\em Molecular ecology}, 19:162--175, 2010.

\bibitem{wyllie1959socialDarwinismAndTheBusinessMan}
I.~G. Wyllie.
\newblock Social {Darwinism} and the businessman.
\newblock {\em Proceedings of the American Philosophical Society},
  103(5):629--635, 1959.

\end{thebibliography}

\end{document}